\pdfoutput=1

\documentclass[11pt]{article}

\usepackage{acl}

\usepackage{times}
\usepackage{latexsym}

\usepackage[T1]{fontenc}

\usepackage[utf8]{inputenc}

\usepackage{microtype}

\usepackage{amsmath}
\usepackage{cleveref}
\crefname{section}{§}{§§}
\usepackage{tabularx}
\usepackage{algpseudocode}
\usepackage{algorithm}
\usepackage{graphicx}
\usepackage{enumitem}

\usepackage[belowskip=0pt,aboveskip=0pt]{caption}
\title{Towards Verifiable Generation: A Benchmark for Knowledge-aware Language Model Attribution}

\author{
 Xinze~Li$^{1}$, Yixin~Cao$^{2\dag}$, Liangming Pan$^{3}$, Yubo~Ma$^{1}$, Aixin Sun$^{1\dag}$ \\ 
 \\
 $^1$ S-Lab, Nanyang Technological University \\
 $^2$ Singapore Management University
 $^3$ University of California, Santa Barbara \\
\texttt{\{xinze002, yubo001\}@e.ntu.edu.sg} \hspace{1cm}
\texttt{axsun@ntu.edu.sg} \\
\texttt{yxcao@smu.edu.sg}\hspace{1cm}
\texttt{liangmingpan@ucsb.edu }\\
}

\begin{document}
\maketitle
\begin{abstract}
Although achieving great success, Large Language Models (LLMs) usually suffer from unreliable hallucinations. Although language attribution can be a potential solution, there are no suitable benchmarks and evaluation metrics to attribute LLMs to structured knowledge. In this paper, we define a new task of Knowledge-aware Language Model Attribution (KaLMA) that improves upon three core concerns with conventional attributed LMs. First, we extend attribution source from unstructured texts to Knowledge Graph (KG), whose rich structures benefit both the attribution performance and working scenarios. Second, we propose a new ``Conscious Incompetence" setting considering the incomplete knowledge repository, where the model identifies the need for supporting knowledge beyond the provided KG. Third, we propose a comprehensive automatic evaluation metric encompassing text quality, citation quality, and text citation alignment. To implement the above innovations, we build a dataset in biography domain BioKaLMA via evolutionary question generation strategy, to control the question complexity and necessary knowledge to the answer. For evaluation, we develop a baseline solution and demonstrate the room for improvement in LLMs' citation generation, emphasizing the importance of incorporating the "Conscious Incompetence" setting, and the critical role of retrieval accuracy.

\end{abstract}

\maketitle
\section{Introduction}

\begin{figure}[t]
\includegraphics[width=\linewidth]{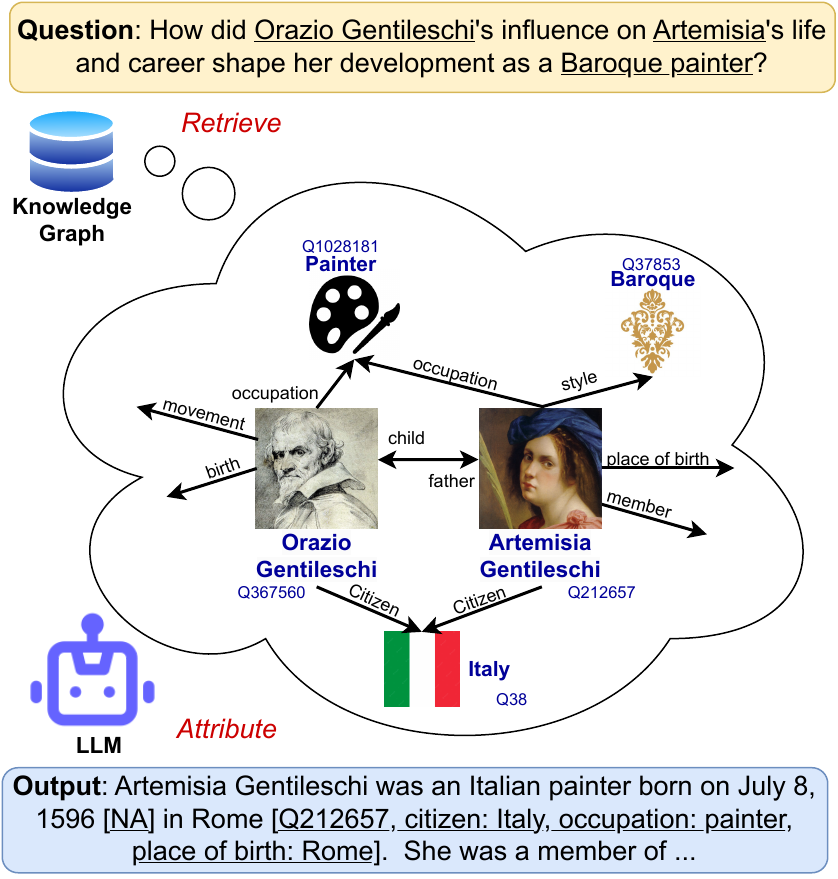}
\caption{A demonstration of our task set up. Given a question, the system generates answers attributed from a retrieved knowledge graph. The underlines in question are the retrieved entities, and the underlines in outputs are the citations. [NA] is the ``Not Applicable Citation''.}
\label{fig:demo}
\end{figure}
Recently, Large Language Models~\cite{brown2020language} (LLMs) have exhibited great capability in open-ended question answering~\cite{yang2019end}. However, the generated answers may include factual errors and are not always reliable, and is commonly known as the ``hallucination''~\cite{shuster2021retrieval, ji2023survey} problem. For instance, LLMs may give wrong diagnosis to patient's symptoms. Hallucination has severe harms especially on industries that require precision and factual knowledge like finance, law, and medical treatment.

To minimize the negative impacts, researchers have proposed the task of language attribution~\cite{bohnet2023attributed}, which not only enables users to verify the generated text flexibly but also contributes to many important applications, such as situation reports ~\cite{reddy2023smartbook}, academic papers~\cite{salvagno2023can}, medical diagnosis~\cite{zuccon2023dr}.Existing works mainly attribute generated outputs to unstructured documents like web pages~\cite{nakano2021webgpt, menick2022teaching} or passages~\cite{gao2023enabling}. To verify the answer quality, they typically compare with a human annotated reference answer for automatic evaluation or conduct human evaluation.
We argue that there are several concerns on such task definition. \textbf{Firstly}, are documents the only source for attribution? Many real-world applications have their own knowledge bases or semi-structured reports.
\textbf{Secondly}, does the attribution source always include all the required knowledge? We consider the coverage issue since no perfect repository can contain all the information in this world.
\textbf{Thirdly}, how to systematically evaluate the attributed content without references? For open-ended questions, there are unlimited number of answers and it is difficult to define a single ground truth. 

To address the first challenge, we utilize knowledge graph (KG) as a reliable source for attribution, namely Knowledge-aware Language Model Attribution (\textbf{KaLMA}). We show a demonstration of task in Figure \ref{fig:demo}. KGs efficiently organize world knowledge in a structured manner and has the potential to unify various formats of data. For example, databases can be easily converted into KGs, or, passages and web pages can be represented as a node in KG like Wikipedia. KaLMA differs from entity linking~\cite{sevgili2022neural} since the sentences or phrases are attributed to a knowledge triplet rather than a single entity. For the second challenge, we tackle the coverage problem by making the model aware of its limitations. We introduce a new setting ``\textbf{Conscious Incompetence}''~\cite{curtiss1974dynamics}, which is the psychological stage that one is aware of the knowledge gap. During generation, LLMs identify sentences that require supporting knowledge absent in the knowledge graph. Our setting enables an attributed LM to recognize the knowledge gaps and allows users to verify uncertain claims, which enhances trustworthiness. For the third challenge, we propose a comprehensive automatic evaluation metric including text quality, citation quality, and text citation alignment. The entire evaluation process does not require human annotated ground truth. 

To implement the above innovations, we first design an automatic dataset construction pipeline. Using this pipeline, we construct a dataset\footnote{The codes and dataset BioKaLMA are publicly available in \url{https://github.com/lixinze777/Knowledge-aware-Language-Model-Attribution}} in the biographical domain, namely \textbf{BioKaLMA}, for a benchmark with all-rounded automatic measurements. Biography forms a good test-set for attribution due to its practical application and convenient evaluation. The availability of high-quality knowledge graph like WikiData also benefits our dataset construction. Derived from the biographical database\footnote{\url{https://plumaj.github.io/biographical/}}~\cite{plum2022biographical} and WikiData, BioKaLMA contains 1,085 data entries. Each data entry includes question and knowledge required to answer the question. For evaluation, we separately evaluate the generated text, the generated citations, and the alignment between texts and citations. We use G-Eval~\cite{liu2023geval} to automatically evaluate the text quality. We also design measurement for correctness, precision, and recall for citations. Lastly, we determine the alignment between texts and citations employing NLI~\cite{dagan2005pascal}

We summarize our contributions as follows: 1) We define the task of  Knowledge-aware Language Model Attribution (KaLMA) that attributes language models to structured knowledge. 2) We design a complete benchmarking pipeline, including dataset, baseline, and evaluation metrics. 3) We conduct extensive experiments and show room for improvement of the LLMs' ability to generate accurate and thorough citations based on provided knowledge graphs. Our experiments on ``Conscious Incompetence'' investigate the capability of current LLMs to identify if there are required knowledge not in knowledge graph. We highlight the necessity of incorporating this setting in future language attribution works. Furthermore, our ablation studies demonstrate the crucial role of retrieval accuracy in achieving desirable generation results.
\section{Task and Dataset}

\subsection{Task Formulation}
We hereby define the task Knowledge-aware Language Model Attribution \textbf{(KaLMA)}: Given a question $q$ and the knowledge graph $G$, the system generates an output text $t$ that answers the question. The output text consists of a list of $m$ sentences $s_1$, ..., $s_m$ grounded with a list of $n$ grounded knowledge $k_1$ .. $k_n$ where $\{k_1 .. k_n\} \in G$. Each knowledge $k$ is a sub-graph of $G$. Each sentence $s$ may be grounded by zero up to multiple knowledge.

\paragraph{\textbf{Setting of Conscious Incompetence}}
We extend this task setting to include conscious incompetence. Given the same input, each sentence $s$ in the output text $t$ can map to a Not Applicable Citation (we use [NA] to represent it) if it includes some knowledge to be verified, but the knowledge is absent in the knowledge graph $G$. A sentence can map to both [NA] and a list of sub-graph knowledge if it can be partially verified by the knowledge graph $G$. [NA] is not a citation on conventional means, but a indicator of knowledge gap.

\subsection{Dataset Construction}
Each entry of dataset bioKaLMA includes two questions and a minimum knowledge set. The two questions enquire about the same people on similar aspects of their life stories. The minimum knowledge set is the smallest set of knowledge that is required to answer each question.
One question is a general version and the other is specific. The general questions are more concise and natural for human readers, and the specific version questions have a tighter bond to the minimum knowledge set, and is hence more accurate for evaluating LLMs. An example data piece is shown in Table \ref{tab:question_example}.

We construct the dataset using an automatic pipeline consisting of three steps: Person Selection, Name Disambiguation, and Evolutionary Question Generation. In the first two steps, we use SPARQL queries to select related people from human written sentences and identify their identity in WikiData. In the third step, we iteratively construct paragraph and question about the selected people. The first iteration starts with a human written sentence about the selected people. In each next iteration, we apply a data selection algorithm to select an appropriate knowledge from WikiData based on the existing paragraph, and extend the paragraph to include the additional knowledge using LLM. Then, LLM constructs the questions using the final paragraph as an answer. The general and specific questions are generated with different prompts and demonstrations. All the selected knowledge from each iteration form the ``minimum knowledge set'' for the question. While we use the human biography domain as an example, this method is applicable to all domains. We present the details of the data construction in Appendix \ref{appendix - dataset construction}.

\begin{table}[t]
\small
    \centering
    \begin{tabular}{>{\raggedright\arraybackslash\tt}p{0.46\textwidth}<{}}
    \hline
        \textbf{General Question}: \\
        \vspace{-1em}
        Who were Oscar and Richard Hertwig, and what were their contributions to the fields of anatomy and biology? \\
        \textbf{Specific Question}: \\
        \vspace{-1em}
        What were the career paths and significant contributions of Oscar and Richard Hertwig in the fields of anatomy and biology, and who were their notable mentors and students? \\
        \\
        \textbf{Minimum Knowledge Set}: \\
        \vspace{-1em}
        \color{brown} {['Q85907', 'occupation', 'biologist']}\\
        \vspace{-1em}
        \color{brown} {['Q85907', 'doctoral student', 'Stanislaus von Prowazek']}\\
        \vspace{-1em}
        \color{brown} {['Q68753', 'doctoral advisor', 'Ernst Haeckel']}\\
        \vspace{-1em}
        \color{brown} {['Q68753', 'student of', 'Ernst Haeckel']}\\
        \vspace{-1em}
        \color{brown} {['Q68753', 'nominated for', 'Nobel Prize in Physiology or Medicine']}\\\hline
  \end{tabular}
  \caption{An example for generated data entry in BioKaLMA. Q85907 and Q68753 are Richard Hertwig and Oscar Hertwig's QIDs in WikiData}
  \label{tab:question_example}
\end{table}

\subsection{Dataset Analysis}
\paragraph{\textbf{Statistics}}
There are 1,085 data entries in BioKalMA. On average, there are 6.8 pieces of knowledge in each ``minimum knowledge set''.
BioKaLMA demonstrates a good demographic variation. It includes a wide range of geographical distribution of people from 196 countries and 949 cities, taking 279 kinds of different occupations. The eras of people span from 1950 B.C. to 2001 A.D. 

\paragraph{\textbf{Evaluation of Dataset Quality}}
We evaluate the BioKaLMA dataset on the following four metrics to ensure the quality of the dataset:
1) \textbf{Authenticity}: The generated questions should accurately reflect the objective facts.
2) \textbf{Relevance}: Each minimum knowledge set should provide support to the corresponding question. Each piece of knowledge from the minimum knowledge set is not redundant.
3) \textbf{Naturalness}: The generated question should be concise and understandable by human readers. 
4) \textbf{Significance}: The generated question should be meaningful and helpful to users.

To our best knowledge, there is no perfect automatic evaluation for these metrics. Naturalness and significance are subjective. Hence, we apply human evaluation to ensure the dataset quality. 

We randomly sample 50 data entries from BioKaLMA and ask human annotators to evaluate the data entries based on the four metrics. The general and specific questions are evaluated separately. More details are given in Appendix \ref{appendix - human evaluation}.

\begin{table}[t]
  {
  \begin{tabular}{l|cc}\hline
     \textbf{Metric (full score)} & \textbf{General} & \textbf{Specific}\\\hline
    Authenticity (1) & 1.00 & 1.00\\
    Relevance (1) & 0.73 & 0.84\\
    Naturalness (5)& 4.38 & 3.52\\
    Significance (5)& 3.94 & 3.68\\\hline
  \end{tabular}
  }
  \caption{Human Evaluation on BioKaLMA dataset.}
  \label{tab:dataset_eval}
\end{table}

The final result for each metric is taken average and reported in Table \ref{tab:dataset_eval}. For both general and specific settings, the questions from sample achieve a 100\% authenticity, which indicates that the overall authenticity of BioKaLMA dataset is high. The relevance on general and specific settings are 73\% and 84\% respectively. The specific question normally consists of more parts and include more details than its general version, and hence some knowledge are necessary to the specific version but not to the general version. However, the general version questions sacrifice relevance to achieve better naturalness and significance. 

In practice, it is difficult to define a precise ``minimum knowledge set'' for a question unless it is very specific. However, a very specific question tends to be artificial. The relevance and naturalness of a question have a trade-off relationship. It is yet challenging to generate questions that have both high relevance and high naturalness, but our generation method allows for a control on the granularity of a question on whether it tends to be more natural or more relevant.


\section{Method} \label{method}
We build a baseline to enable LLMs to generate knowledge-aware attributed answers. Following the approach of many retrieval augmented generation works~\cite{lee2022need, izacard-grave-2021-leveraging}, we utilize a pipeline consisting of three components: retrieval, re-ranking, and generation.

\subsection{Retrieval}
Our baseline retrieval process consists of two parts: named entity recognition and graph retrieval. We utilize spaCy\footnote{\url{https://spacy.io/api/entityrecognizer}} to identify the named entities mentioned in the question. Using these entities, we retrieve entity-centered sub-graphs using SPARQL. For each retrieved entity, we search for nodes in the graph that match the entity's name. We use the named entity recognition (NER) entity type as a simple filter (e.g., the NER category ``person'' matches the ``human'' entity type in WikiData). Taking each selected node as the center, we retrieve one-hop sub-graphs that contain properties associated with the entity.

\subsection{Re-ranking}
The re-ranking component plays a crucial role in disambiguating retrieved entities, as multiple entities may share the same name in the WikiData graph. Two common scenarios are different individuals with the same name (e.g., Anne Hathaway the American actress and Anne Hathaway the wife of William Shakespeare) and different references to the same word (e.g., ``Chinese'' the language and ``Chinese'' the ethnic group). When multiple entities are retrieved from the graph for a given entity name, we rank the graphs based on the Exact Match (EM) between the neighboring nodes and the question. We select the entity with the highest number of matched neighboring nodes.  

\subsection{Generation}

The generation component effectively prompt the LLMs with the retrieved knowledge graphs (KGs) to generate answers that attribute the KG. To adapt to the input format of the LLMs, we transform the structured KGs into flat texts. We preserve the information of the retrieved sub-graphs by mapping each sub-graph to a set of triples. Each triple consists of two nodes and one edge, where one node is the centered entity, the other node is its neighbor, and the edge represents the relationship between them. For example, [Q212657 - place of birth - Q220] can be translated to [Artemisia Gentileschi - place of birth - Rome]. In this translation, we use the names of the entities for better comprehension by both the models and humans, since WikiData utilizes QIDs (e.g., Q220) to represent unique entities.
We construct a prompt (Table \ref{tab:main_prompt} in appendix \ref{appendix - prompts}) which includes 1) instruction to the models to generate attributed answers. 2) retrieved knowledge graph,  and 3) the question. We employ one-shot in-context learning~\cite{brown2020language} by prepending one human written demonstration. 
In the one-shot demonstration, we use the special token [NA] to represent the ``Not Applicable Citations'' for conscious incompetence. We deliberately omit some knowledge in the demonstration example knowledge graph, and we insert [NA] tokens in the corresponding sentences that use these knowledge within the example answer.


\section{Evaluation Metrics}
Our benchmark includes evaluation metrics for both the generated text and citations. We also evaluate the alignment between the text and corresponding citations. We provide more discussions on the design of evaluation metrics in subsection \ref{dis - evaluation metrics}.
\subsection{Text Evaluation}
Since our test-set has no human-written gold answers as references, we do not utilize comparison-based metrics such as BERTScore~\cite{zhang2019bertscore} or MAUVE~\cite{pillutla2021mauve}. Instead, we employ reference-free NLG evaluator G-Eval~\cite{liu2023geval}, which defines the following four metrics: 1) \textbf{Coherence}: whether the generated text is well-structured and well-organized. 2) \textbf{Consistency}: whether the generated text is consistent with the knowledge provided. 3) \textbf{Fluency}: whether the generated text is well-written and grammatical. 4) \textbf{Relevance}: how well is the generated text relevant to the question.

We use the model text-davinci-003 for evaluation, which assigns an integer score of 1 to 5 for each metric. We follow the prompt provided in G-Eval~\cite{liu2023geval} and customize it based on our task. The full prompts are given in \cref{appendix - prompts}. 

\begin{figure}[t]
\includegraphics[width=\linewidth]{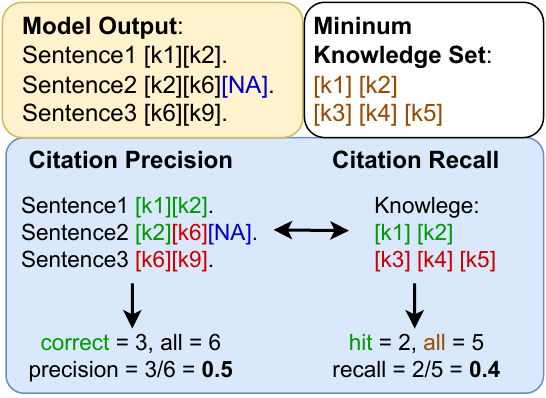}
\caption{An illustration of how we evaluate the precision and recall for generated citations.}
\label{fig:citation_precision_recall}
\end{figure}

\subsection{Citation Evaluation}
We evaluate the citation qualities from three aspects: 1) \textbf{Correctness}, which measures whether the generated knowledge matches the given knowledge from the knowledge graph, 
2) \textbf{Precision}, which determines how much of the generated citations are helpful to answer the question, and
3) \textbf{Recall}, which measures how much of the minimum knowledge set are covered by the generated citations.
We also calculate the F1-Score based on the Precision and Recall to reflect the overall quality of citations. 

\paragraph{\textbf{Correctness}} 
We calculate the citation correctness for each citation (0 or 1) and average over all citations. Each citation comprises a triplet of 1) center entity QID, 2) relation 3) neighbour entity value. If the generated citation is complete with all three parts, and exactly matches a triplet from the question's retrieved KG, correctness = 1.

\paragraph{\textbf{Precision}}
We calculate citation precision for each citation (0 or 1) and average over all citations to get micro precision. Precision = 1 for a citation if and only if 1) it is correct, and 2) it matches one knowledge triplet from minimum knowledge set of the question.  (See Figure \ref{fig:citation_precision_recall}.)

\paragraph{\textbf{Recall}}
We calculate citation recall for each knowledge (0 or 1) in minimum knowledge set, and average over all knowledge to get micro recall. Recall = 1 if and only if the knowledge if hit by a correct citation. 
 (See Figure \ref{fig:citation_precision_recall}.)

We average over all citations/knowledge in an answer, and average all answer-level precision/recall to get macro precision and recall. we calculate micro and macro F1-Score from corresponding precision and recall.

\begin{figure}[t]
\includegraphics[width=\linewidth]{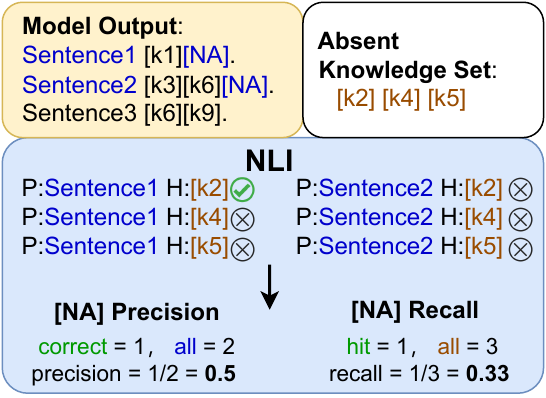}
\caption{An illustration of how we evaluate the precision and recall for conscious incompetence ([NA])}
\label{fig:na_precision_recall}
\end{figure}

\subsection{Text-Citation Alignment}
Other than the text quality and citation quality, we measure whether the generated citations provide support for the corresponding sentences. A piece of useful knowledge is not an ideal citation if it is irrelevant to the sentence it links to. Therefore, we propose the metric ``Alignment'' which determines whether the generated citations are aligned to the sentences to which they belong. We use a state-of-the-art natural language inference (NLI) model TRUE~\cite{honovich2022true}, which is a fine-tuned T5-11B~\cite{raffel2020exploring} model, to check whether the generated sentence entails the generated citation. Since one sentence could have multiple citations, we run NLI on all sentence-citation pairs and report the percentage of entailment. Additionally, we conduct human evaluation in \cref{human_eval} to showcase if the automatic evaluation is correlated with human judgments.

\subsection{Conscious Incompetence Evaluation} \label{Conscious Incompetence Evaluation}

Theoretically, each [NA] mark should map to a piece of knowledge absent from the retrieved knowledge graph. However, it is difficult to identify if sentence requires any absent knowledge since there is no ground truth. Therefore, we conduct a three-round experiment to manually create ground truth for absent knowledge. 
In round 1, we select one knowledge from the minimum knowledge set, and remove it from the ground-truth knowledge graph. We let the LLMs attribute to this incomplete knowledge graph to generate answers, whereby the removed knowledge forms the ``absent knowledge ground truth''. In subsequent rounds, we each remove one additional knowledge from the minimum knowledge set, simulating a knowledge graph with more serious coverage problem.

We employ the NLI model TRUE~\cite{honovich2022true} to measure the alignment between sentences and knowledge. A sentence with [NA] should be aligned to an absent knowledge. We calculate precision and recall for [NA].

\paragraph{\textbf{[NA] precision}} 
We calculate [NA] precision for each sentence with [NA] (0 or 1) and average over all sentences with [NA]. Precision = 1 for a sentence if and only if it entails one knowledge triplet from absent knowledge set of the question.  (See Figure \ref{fig:na_precision_recall}.)

\paragraph{\textbf{[NA] Recall}}
We calculate [NA] recall for each knowledge (0 or 1) in absent knowledge set and average over all absent knowledge. Recall = 1 if and only if the knowledge if entailed by a sentence with [NA]. 
 (See Figure \ref{fig:na_precision_recall}.)

\subsection{Discussion on Evaluation metrics} \label{dis - evaluation metrics}
In this section, we discuss on the evaluation metrics of benchmark BioBaLMA. We design the evaluation metrics from multiple dimensions to incorporate different understandings on what makes a high quality citation. 
\begin{itemize}
    \item One understanding argues when the answer contains mistakes, even if the citation is correctly answering the questions, it cannot represent good LLM attribution performance. In this case, citation quality is considered as a measure of overall attribution performance, including the answer quality.
    \item The other understanding argues for a complete decoupling of answer and citation quality. In this scenario, even if the answer is wrong, the citation is valuable as long as it provides reasonable support for the question. In such cases, citations do not give advice on the correctness of the answer.
\end{itemize}
Both understandings are plausible, and hence we have considered both of them when we design metrics. The alignment score is designed based on the first understanding, which measures whether the citations are closely linked to the answer. The precision and recall are designed for the second understanding, where the citations are completely decoupled from the answer, and are correct if they provide support for the question. 

In addition, we also incorporate an edge case for design of the [NA] precision calculation. If an NA-marked sentence does not answer the question at all, it is considered correct in the [NA] precision calculation. In this case, the LLM correctly identifies a sentence that requires further verification.  
\begin{table*}[t]
    \centering
  {
  \begin{tabular}{l|c|cccc|ccc}\hline
     &\multicolumn{5}{c}{\underline{\textbf{Micro}}} & \multicolumn{3}{c}{\underline{\textbf{Macro}}}\\
     \textbf{Model} & \textbf{Align.} & \textbf{Corr.} & \textbf{Prec.} & \textbf{Rec.} & \textbf{F1.} & \textbf{Prec.} & \textbf{Rec.} & \textbf{F1.} \\\hline
    GPT-4 (0.5) & \textbf{92.0}$_{(1.5)}$ &\textbf{97.6}$_{(0.1)}$ & \textbf{36.0}$_{(0.6)}$ & 43.6$_{(1.0)}$ & \textbf{39.4} & \textbf{40.7}$_{(1.1)}$ & 43.9$_{(1.0)}$ & \textbf{42.3}\\
    ChatGPT (0.1) & 85.9$_{(2.5)}$ & 96.1$_{(0.4)}$ & 29.0$_{(0.0)}$ & \textbf{50.8}$_{(0.3)}$ & 36.9 & 32.7$_{(0.4)}$ & \textbf{51.2}$_{(0.3)}$ & 39.9\\
    ChatGPT (0.5) & 84.5$_{(1.1)}$ & 94.8$_{(0.2)}$ & 29.9$_{(0.2)}$ & 49.0$_{(0.8)}$ & 37.2 & 34.1$_{(0.5)}$ & 49.4$_{(0.9)}$ & 40.4\\
    ChatGPT (0.9) & 84.1$_{(0.5)}$ & 94.2$_{(0.4)}$ & 28.7$_{(0.2)}$ & 49.0$_{(0.3)}$ & 36.2 & 32.5$_{(0.2)}$ & 49.4$_{(0.3)}$ & 39.2\\\hline
    Alpaca-7B  & 46.9$_{(0.9)}$ & 78.9$_{(0.6)}$ & 14.9$_{(1.4)}$ & 19.4$_{(0.2)}$ & 16.8 & 19.8$_{(0.4)}$ & 19.9$_{(0.3)}$ & 19.8\\
    LLaMA-7B  & 47.8$_{(0.8)}$ & 70.2$_{(0.2)}$ & 7.7$_{(2.4)}$ & 41.1$_{(0.7)}$ & 13.0 & 11.0$_{(1.9)}$ & 41.4$_{(0.7)}$ & 17.4\\
    LLaMA-13B & 62.1$_{(0.4)}$ & 71.7$_{(1.9)}$ & 10.5$_{(3.3)}$ & 43.7$_{(1.0)}$ & 16.9 & 13.8$_{(2.2)}$ & 43.5$_{(1.0)}$ & 20.9\\
    Vicuna-13B  & 66.9$_{(0.1)}$ & 59.0$_{(0.6)}$ & 14.9$_{(0.2)}$ & 16.8$_{(0.0)}$ & 15.8 & 15.1$_{(0.0)}$ & 17.0$_{(0.0)}$ & 16.0\\\hline
  \end{tabular}
  }
  \caption{Citation Quality OpenAI models and LLaMA family models. The first five metrics are reported in Micro, and the last three metrics are reported in Macro. We also report text citation alignment.}
  \label{tab:main_result}
\end{table*}



\section{Experiments}
We run through the method pipeline described in \cref{method} on different LLMs and present the results in this section. Since we aim to obtain a more accurate evaluation, we conduct our main experiments on the specific questions setting, since the minimum knowledge set has a higher relevance on the specific questions. However, we will also provide evaluation results for the general questions in \cref{general_questions} as ablation studies.
The implementation details are reported in \cref{appendix - implementation details}. We report five model baselines from both open and closed source model families:

\paragraph{\textbf{OpenAI Models}} 
We use GPT4 (gpt-4-0314) and ChatGPT (gpt-3.5-turbo-0301) for our experiments. For ChatGPT, we experiment on temperature of 0.1, 0.5, and 0.9 to obtain different levels of randomness and creativity in generation.

\paragraph{\textbf{LLaMA}}
We conduct experiments with LLaMA-7B~\cite{touvron2023llama} and LLaMA-13B since they are powerful open-source models that are widely accessible. We have also conducted human instruction tuned LLaMA models, including Alpaca-7B~\cite{taori2023alpaca} and Vicuna-13B~\cite{vicuna2023}.

\subsection{Main Results}
\paragraph{\textbf{Citation Quality Evaluation}}
We present the main results in Table \ref{tab:main_result}. For correctness, we report on a micro scale. For precision, recall, and F1-Score, we report on both micro and macro scales. The experimental results are the mean of three runs, and the standard deviation is reported in brackets.

In general, there is a room of improvement for all models since no model can achieve a micro F1 Score of higher than 40.  
The OpenAI models outperform the LLaMA family models in almost all metrics. The correctness is above 94 for OpenAI models, but around 70 for LLaMA based models.
For ChatGPT, temperature does not play a significant role since it effect on F1 Score is at most 1.2.
The GPT-4 model achieves the best performance across almost all metrics, except for recall, since GPT-4 models tend to generate shorter answers with fewer citations, resulting in higher precision. 
While LLaMA is better at Recall by generating long answers with many citations. The F1-Score of models from the same family are close to one another, showing that our automatic evaluation metric designed is reliable.

\paragraph{\textbf{Text-Citation Alignment}} From Table \ref{tab:main_result}, similar to citation quality, the OpenAI models also outperform the LLaMA based models on text-citation alignment. 
In addition, models with 7B, 13B, 175B (ChatGPT), and trillion level (GPT4) parameters have an alignment score of 40+, 60+, 80+, and 92 respectively. LLaMA-13B model has an improvement of 14.3 compared to LLaMA-7B model. This shows that parameter size may play an important role in generating sentences and citations with good alignment. 

\paragraph{\textbf{Text Quality Evaluation}}
\begin{table}[t]
  {
  \begin{tabular}{l|cccc}\hline
     \textbf{Model} & \textbf{Coh.} & \textbf{Con.} & \textbf{Flu.} & \textbf{Rel.}\\\hline
    GPT-4 (0.5) & 4.48 & 4.89 & 4.64 & 4.72\\
    ChatGPT (0.1) & \textbf{4.57} & \textbf{4.94} & 4.69 & \textbf{4.83}\\
    ChatGPT (0.5) & \textbf{4.57} & \textbf{4.94} & \textbf{4.71} & 4.81\\
    ChatGPT (0.9) & 4.52 & 4.91 & 4.67 & 4.79\\\hline
    Alpaca-7B  & 4.10 & 4.46 & 4.23 & 3.76\\
    LLaMa-7B  & 3.06 & 3.79 & 3.62 & 2.96\\
    LLaMa-13B & 3.60 & 4.23 & 3.94 & 3.56\\
    Vicuna-13B  & 3.67 & 4.50 & 3.96 & 3.64\\\hline
  \end{tabular}
  }
  \caption{Evaluation on generated text quality.}
  \label{tab:text_eval}
\end{table}

We present the evaluation of generated text quality in Table \ref{tab:text_eval}. From the results, we find that OpenAI models, in general, have better text quality in all metrics compared to LLaMA family models, which corresponds to the citation evaluation results. All models exhibit rather high consistency, indicating that the LLMs are capable of generating answers that are not contradictory to the provided knowledge or self-contradictory. However, the relevance is relatively low for smaller models, indicating the difficulty these models face in generating answers that are relevant to the questions.

\subsection{Conscious Incompetence} \label{NA}
\begin{table}[t]
  {
  \begin{tabular}{l|cccc}\hline
     \textbf{Removed} & \textbf{Corr.} & \textbf{Prec.} & \textbf{Rec.} & \textbf{F1.}\\\hline
    0 (gold) & 95.5 & 30.1 & 57.1 & 39.4\\
    1 & 94.1 & 26.1 & 42.5 & 32.3\\
    2 & 94.0 & 21.0 & 31.4 & 25.2\\
    3 & 93.9 & 16.3 & 20.4 & 18.1\\\hline
  \end{tabular}
  }
  \caption{Citation quality evaluation for generated texts using a KG with N pieces of knowledge removed.}
  \label{tab:NA_eval}
\end{table}

We first evaluate \textbf{citation quality} of the generated text with knowledge removed using method described in \cref{Conscious Incompetence Evaluation}. From Table \ref{tab:NA_eval}, the removal of required knowledge has a minimal impact on correctness, but significantly affects citation precision and recall. With more knowledge absent from provided knowledge graph, both precision and recall drops drastically, demonstrating that the coverage issue poses a considerable challenge to generating answers with high quality citations.

\begin{figure}[t]
\includegraphics[width=\linewidth]{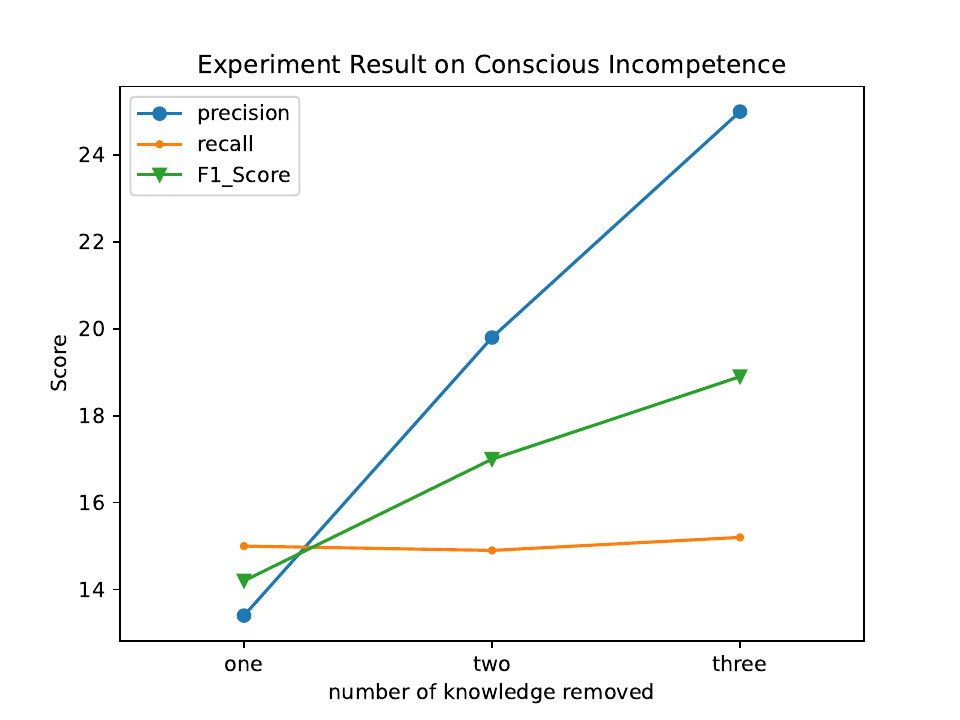}
\caption{Precision, Recall, and F1-Score for [NA].}
\label{fig:NA_P_R}
\end{figure}

Next, we evaluate \textbf{[NA] precision} and \textbf{recall}. From Figure \ref{fig:NA_P_R}, The recall is stable at about 15 regardless of the number of absent knowledge. This indicates that the current LLMs have ability to identify absent knowledge to a limited extent.
While precision and F1-Score exhibit a clear upward trend, which shows that with more absent knowledge in KG, [NA] enables generated outputs to locate absent knowledge more accurately. Therefore, the ``Conscious Incompetence'' setting plays an increasingly crucial role when the coverage problem of knowledge graph is more serious.



\subsection{Retrieval Analysis}
\begin{figure}[t]
\includegraphics[width=\linewidth]{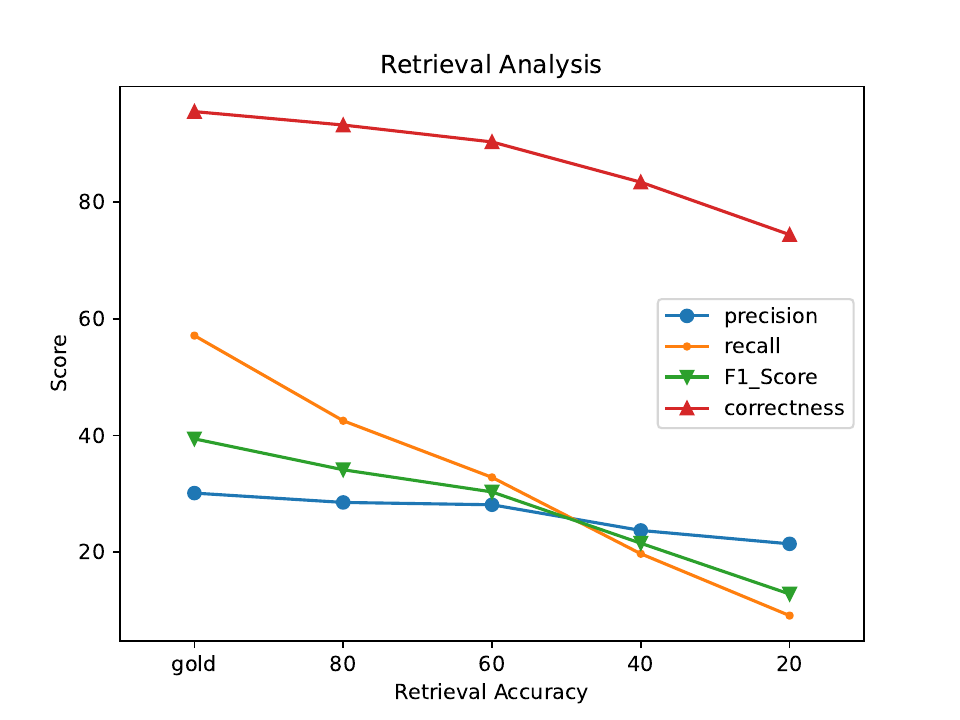}
\caption{Citation evaluation (Micro) of generated texts using knowledge graphs with retrieval accuracy 100 (gold), 80, 60,40, and 20.}
\label{fig:retrieval_ablation}
\end{figure}

We conduct an ablation study to examine the impact of retrieval accuracy on the model's output. The experiment simulates retrieval accuracy from 100 to 20 at intervals of 20. We start with the ground truth knowledge graphs that we used for question construction. In each subsequent rounds, we randomly replace additional 20\% knowledge graphs with irrelevant knowledge graphs to simulate retrieving wrong graphs. The results for citation quality are in Figure \ref{fig:retrieval_ablation}. Answers are generated using ChatGPT with a temperature of 0.5.

The results show clear downward trends in all metrics as expected when retrieval accuracy dropped. Among precision and recall, the impact of poor retrieval quality on recall (green) is much more significant than on precision (yellow). This indicates that the model has the ability to filter out incorrect knowledge to a certain extent, resulting in less noticeable impact on precision compared to recall. The reduction in recall was nearly linear as retrieval accuracy decreased, which is understandable since a knowledge cannot be cited if it is not provided. The greatest drop in recall occurred between the ground truth (57.1) and 80 accuracy (42.5), demonstrating the potential of the model to generate high-quality citations under perfect retrieval conditions. In practice, a retrieval accuracy of 80 is closest to the actual scenario of our experiment (our retrieval accuracy is 75.9). Therefore, when retrieval accuracy is reasonably high, the correctness of citations is not the most significant concern compared to recall.


\begin{table}[t]
  {
  \begin{tabular}{l|cc}\hline
      & \textbf{Alignment} & \textbf{Human Avg.}\\\hline
   ChatGPT(0.5) & 84.5 & 82.0 \\
   LLaMA-7B & 47.8 & 45.5 \\
   Vicuna-13B & 66.9 & 64.5 \\\hline
  \end{tabular}
  }
  \caption{Result of Human Evaluation on text-citation alignment}
      \label{tab: human_eval}
\end{table}

\begin{table*}[t]
    \centering
  {
  \begin{tabular}{l|l|c|cccc|cccc}\hline
     &\multicolumn{7}{c}{\underline{\textbf{Citation Eval.}}} & \multicolumn{2}{c}{\underline{\textbf{Text Eval.}}}\\
     \textbf{Setting} &\textbf{Model} & \textbf{Align.} & \textbf{Corr.} & \textbf{Prec.} & \textbf{Rec.} & \textbf{F1.} & \textbf{Coh.} & \textbf{Con.} & \textbf{Flu.} & \textbf{Rel.} \\\hline
    General & GPT-4 (0.5) & 90.9 &\textbf{97.6} & 30.8 & 42.1 & 35.6 & 4.38 & 4.77 & 4.48 & 4.48\\
    & ChatGPT (0.5) & 82.7 & 94.5 & 25.2 & 47.4 & 32.9 & \textbf{4.64} & 4.89 & 4.45 & 4.70\\\hline
    Specific & GPT-4 (0.5) & \textbf{92.0} &\textbf{97.6} & \textbf{36.0} & 43.6 & \textbf{39.4} & 4.48 & 4.89 & 4.64 & 4.72\\
    & ChatGPT (0.5) & 84.5 & 94.8 & 29.9 & \textbf{49.0} & 37.2 & 4.57 & \textbf{4.94} & \textbf{4.71} & \textbf{4.81} \\\hline
  \end{tabular}
  }
  \caption{Comparison of evaluation results on General and Specific question setting}
  \label{tab:main_result_general}
\end{table*}

\subsection{Human Evaluation}\label{human_eval}

We conduct human evaluation to verify the correlation between automatic evaluation and human judgment. We randomly sample 100 sentence-citation pairs from each of the three baselines: ChatGPT (temperature 0.5), LLaMA-7B, and Vicuna-13B. We request two proficient English annotators for each baseline to determine if the citation aligns to the sentence and provides support for it. The reason we choose metric alignment here is in \cref{appendix - human evaluation}, with instruction to annotators and IAA.

The comparison between automatically calculated Alignment and human evaluation results is shown in Table \ref{tab: human_eval}. For all three baselines, the automatic and human scores are close with a gap within 2.5, despite the significant differences among the baselines. This indicates a strong correlation between the automatically calculated alignment and human judgments. The experiment results demonstrate that the automatic evaluation serves as a reliable measurement of the alignment between generated texts and citations.

\subsection{General and Specific Questions} \label{general_questions}
We compare experiments results of text, citation (micro), and alignment between the general and specific questions in Table \ref{tab:main_result_general}. The results show that the same model's answers on specific questions outperform those on general questions in almost all metrics. The finding is not surprising because the specific questions provide clearer instructions to the models on which knowledge to use. In addition, the general questions in the dataset are inherently loosely bonded to the minimum knowledge set, and hence have impacts on the evaluation results. This experiment shows a trade-off between how explicitly the question context mentions the knowledge, and how irreplaceably the knowledge is required by the question. The specific questions target the knowledge more explicitly in the question context, and hence cover the scope of the paragraph better. It stands for an upper bound for knowledge coverage and a lower bound for question naturalness.The general questions implicitly target the knowledge in the question context, and there loosely cover the scope of the paragraph. It stands for an upper bound for question naturalness and a lower bound for knowledge coverage.



\section{Related Work}\label{related work}
\paragraph{\textbf{Retrieval-augmented LLMs}} 
KiC~\cite{pan2022knowledge} empower models with external memory of multiple formats including knowledge graph but does not explore attribution.
WebGPT~\cite{nakano2021webgpt} outsources document retrieval to Microsoft Bing and fine-tunes GPT3 to answer questions. 
GopherCite~\cite{menick2022teaching} fine-tunes a Gopher~\cite{rae2021scaling} model to generate text alongside quotes extracted from Google search.
ALCE~\cite{gao2023enabling} retrieves top-k passages from Wikipedia and asks LLMs to generate outputs with citations to corresponding supporting documents. 
These works attribute LLMs to unstructured documents but not knowledge graph.

\paragraph{\textbf{Evaluation}}
\cite{rashkin2021measuring} define the ``Attributable to Identified Sources'' (AIS) to measure whether model-generated statements are supported by underlying sources.
\cite{bohnet2022attributed} study an automatic metric (AutoAIS) that formulates evaluation of automated question answering as a NLI task.
\cite{yue2023automatic} investigate the automatic evaluation of attribution by prompting LLMs and fine-tuning smaller LMs.
\cite{liu2023evaluating} conduct human evaluation to audit generative search engines for their citation qualities.
ALCE~\cite{gao2023enabling} evaluates generated answers by comparing with gold answers using MAUVE, and calculates precision and recall for citations using NLI.
To the best of our knowledge, our evaluation methods are the first framework that requires no human annotated data.
\section{Conclusion}

We propose KaLMA that comprises a new dataset BioKaLMA, a pipeline for generating attributed answers by retrieving from KGs, and a set of automatic evaluation metrics to assess text quality, citation quality, and text-citation alignment. We introduce the ``Conscious Incompetence'' setting, enabling LLMs to identify the knowledge required to support the answers but is absent from the KG. Through this benchmark, we address three challenges: incorporating diverse attribution sources, limited attribution source coverage, and the absence of human annotated ground truth for automatic evaluation.

Our extensive experimental results demonstrate that current LLMs still have room for improvement when utilizing KGs as attribution sources. We also highlight the increasing effectiveness of ``Conscious Incompetence'' setting as the coverage of attribution source becomes worse. Lastly, we prove the crucial role of retrieval accuracy in generating high-quality attributed texts.

\section*{Limitations}
One limitation is that our work only investigates a simple form of knowledge graph, where each node is an entity, and each sub-graph is a knowledge triple. There are more complicated forms of knowledge graph, where each node is a document. We will explore this setting in future works.

Another limitation lies within the text quality evaluation. We uses ChatGPT as the model to evaluate texts, which could potentially have a bias if the model prefers the text style generated by itself. Such bias can be observed from the abnormal phenomenon that the scores of ChatGPT generated answers are higher than that of the GPT4 generated answers for all four dimensions. Due to cost considerations, we do not repeat the text quality evaluation with GPT-4.

\section*{Ethical Considerations}
The potential risk is when users leverage the automatic dataset construction pipeline to generate massive hazardous datasets. This can only happen when a structured knowledge of harmful content is available. Otherwise there is no risk as long as the benchmark is used correctly. All data are collected from WikiData which is publicly available. Hence there is no privacy issue. We also conduct human check to ensure there is no offensive content.

\bibliography{anthology,main}
\appendix
\newpage

\section{Dataset Construction} \label{appendix - dataset construction}
In this section, we will explain the detailed process and algorithms for the automatic dataset construction pipeline. Using this pipeline, we are able to construct datasets with a greater scale or in other domains.
\subsection{Person Selection}
To improve the complexity of the questions and difficulty to LLMs, we involve more than one person in each question. In addition, we need high quality paragraphs for subsequent dataset generation steps. Therefore, we utilize name pairs and paragraphs from the \textbf{biographical database}, which is a database specifically designed for the relation extraction (RE) task. Each piece of data from the biographical database includes a short paragraph, and a relation triple extracted from the paragraph. The relation triple consists of two people and their relationship such as <William Shakespeare, Spouse, Anne Hathaway>.  The biographical database includes an automatically extracted set and a human annotated set. We specifically choose the human annotated set from the database to ensure high-quality name pairs. To avoid potential ambiguities, we filter out data if any name in the triple is incomplete. In practice, we consider a name complete if it has at least a family name and a surname.

\subsection{Name Disambiguation}
Due to the presence of duplicate names (e.g., Anne Hathaway: the actress, or the wife of William Shakespeare), we perform name disambiguation to map each name in the triple to a unique entity from the knowledge graph. We utilize WikiData\footnote{\url{https://www.wikidata.org/wiki/Wikidata:Main_Page}}~\cite{vrandevcic2014wikidata} as the knowledge base and employ SPARQL~\cite{perez2009semantics} queries to retrieve all entities associated with the name. WikiData assigns a unique QID to each entity which distinguishes between entities with the same name. In WikiData, each entity represents a node in the knowledge graph. Since each triple consists of two names and one relation, we select the two entities obtained from the query if they are connected to each other on WikiData. Additionally, the connecting edge should align with the relation specified in the triple. Subsequently, we extract the one-hop sub-graph centered around each person node, which provides properties related to the person, such as gender, birth date, occupation, and more. We convert ambiguous person names from previous steps to unique QID from WikiData. The extracted sub-graphs contain all knowledge from WikiData about the selected people. We call the extracted graphs ``knowledge pool''.

\subsection{Evolutionary Question Generation}
We employ an ``evolutionary question generation'' approach inspired by WizardLM~\cite{xu2023wizardlm} and DKGen~\cite{qian2023optimizing}, where we gradually increase the set of knowledge required by injecting knowledge through iterations. In each iteration, LLMs extend the paragraph with one sentence by incorporating the additional knowledge. After the last iteration, LLMs propose two questions according to the extended paragraph, one is a general version, and the other is a specific version. The general question is more concise, and the specific question is more detailed. Both questions target the same set of knowledge. All injected knowledge form a ``\textbf{minimum knowledge set}'', which includes the least knowledge required to answer the proposed question (Table \ref{tab:question_example}). We do not throw all knowledge to LLM at once to form a paragraph because extending the paragraph and knowledge set incrementally allow us to select the appropriate knowledge after each iteration. 

In the first iteration, LLMs annotate the original paragraph from Biographical Database with the knowledge from the ``knowledge pool''. For instance, the sentence ``Artemisia was born in Rome.'' is annotated with  knowledge [Artemisia , place of birth, Rome].
In each subsequent iteration, we select a piece of appropriate knowledge according to the existing paragraph. A sentence with appropriate knowledge should have good \textbf{specificity} and \textbf{coherence}. Specificity refers to the significance of the knowledge, such that it is not too general or trivial. Coherence refers to the naturalness of the additional knowledge. The added knowledge should not deviate from the existing paragraph and should be coherent when reading.
During knowledge selection, each piece of knowledge is assigned a score by adding the specificity score and coherence score.
The specificity score measures the uniqueness of the knowledge. We discourage the system from selecting too frequent relation types like ``gender'' or "date of birth" which may be less informative. A less frequent relation tend to provide a knowledge specific to the person. Derived from IDF, we calculate the \textbf{number of occurrences $Count_r$} for each \textbf{relation $r$} in the \textbf{dataset with size $N$}. 
The coherence score is calculated through perplexity. We convert each piece of knowledge to a simple sentence by applying a template. For instance, the knowledge [Artemisia, place of birth, Rome] is converted to ``Artemisia's place of birth is Rome''. There are three templates depending on the POS of the relation. We append each sentence to the original paragraph and calculate normalized inverse perplexity to obtain coherence score. 
The overall score is a weighted sum of specificity score and coherence score:
\begin{equation}
    \begin{split}
    Score_r = &\, \alpha \cdot \log(2 \cdot N / Count_r) \\
              &+ (1-\alpha) \cdot \text{softmax}(1 / perp_r)
    \end{split}
\end{equation}
In each iteration, we leverage the ``text-davinci-003'' model for annotation or generation with in-context learning. We provide separate instructions and demonstrations for general and specific questions. The detailed prompt templates used is provided in the 
\cref{appendix - prompts}.
We provide one human written demonstration. Some examples of full question evolution process are provided in 
\cref{appendix - evolutionary}.
In practice, we employ five iterations to ensure sufficient complexity in the questions without making them overly tedious.

\section{Experiment Details}\label{appendix - implementation details}
\subsection{Main Experiment}
For the main experiments, we run reach model with different seeds for three times. The OpenAI family models are implemented using OpenAI APIs. Running one round of experiment with ChatGPT model takes approximately 1 hour, and costs about 3 USD. Running one round of experiment with GPT4 model takes approximately 1.5 to 2 hours, and costs about 60 USD. Each LLaMA family model is run on one TESLA V100 GPU, where each run takes about 6 to 8 hours for Alpaca-7B and Vicuna-13B, and about 12-16 hours for LLaMA-7B and LLaMA-13B.

\subsection{Text Quality Evaluation}
For text quality evaluation, we use the model ``text-davinci-003'' with temperature 0 to ensure stability and reproducibility of the results. We randomly sample 100 outputs from each baseline and take three runs to report mean. We do not report standard deviation since most of them are mostly insignificantly small (below 0.1). 

\subsection{NLI}
For the automatic evaluation of text citation alignment and evaluation of the known unknown citations, we implement the TRUE model from HuggingFace\footnote{ \url{https://huggingface.co/google/t5_xxl_true_nli_mixture}}, which was trained on 
SNLI~\cite{bowman2015large_snli},
MNLI~\cite{williams2018broad_mnli},
Fever~\cite{thorne-etal-2018-fever},
Scitail~\cite{khot2018scitail},
PAWS~\cite{zhang-etal-2019-paws},
and VitaminC~\cite{schuster-etal-2021-get}.
The model uses the prompt of ``premise: \{PREMISE\} hypothesis: \{HYPOTHESIS\}''. For each sentence citation pair, we place the sentence in the ``PREMISE'', and the citation to the ``HYPOTHESIS'', like the following: ``premise: \{Hertwig served as a professor at the University of Jena for the last 40 years of his career.\} hypothesis: \{employer: University of Jena\}''

\section{Human Evaluation}\label{appendix - human evaluation}
\subsection{Dataset Evaluation}
To evaluate the dataset quality, we have two individual annotators who are proficient in the English language. 
Below are the exact method for evaluating each metric:
\begin{itemize}
    \item \textbf{Authenticity}. We ask the annotators to check from WikiPedia and understand the background stories of the mentioned people, and decide if the generated question matches the background story. Each question is assigned score 1 if it matches the background story, and score 0 if there is contradiction. 
    \item \textbf{Relevance}. After understanding the background stories, we ask the annotators to label each piece of knowledge from the minimum knowledge set. A piece of knowledge is labeled 1 if the annotator thinks it is necessary to answer the question, and 0 if it is redundant. The relevance score of a question is the ratio of number of necessary knowledge to the number of knowledge in the minimum set. 
    \item \textbf{Naturalness}. We ask the annotators to give an integer score 1 to 5 to label each question. 5 means the question can be easily understandable, and is concise. 1 means the question is not written in natural English language or is extremely tedious.
    \item \textbf{Significance}. We ask the annotators to give an integer score 1 to 5 to label each question. 5 means the annotator feels that he or she may be interested in this question under some circumstances, and 1 means the opposite.
\end{itemize} 
The agreement between the two annotators are as follow: the agreement between them is 100\% for authenticity and 86\% for relevance. Since the evaluation for naturalness and significance are score based, in 92\% and 90\% of the evaluated datasets respectively, the score difference between the two annotators is no larger than 1. 

\subsection{Generated Text Evaluation}
Among text quality evaluation, citation quality evaluation, and text-citation alignment, we conduct human evaluation on text-citation alignment.
Text quality evaluation is conducted using G-Eval. We acknowledge this is not a perfect metric, but the human evaluation is conducted in~\cite{liu2023geval}. The focus is this paper is not to improve G-Eval.
Citation quality evaluation is conducted with looking for exact match between generated citations and minimum knowledge set, which is an objective evaluation.
The text-citation alignment evaluation is conducted using NLI, which we are not certain if entailment means providing support. In addition, whether a knowledge supports a sentence can be subjective. Therefore, we conduct human evaluation on alignment.

We present the Human Evaluation Instructions provided to the annotators in Table \ref{tab:human_eval_template}. We follow the implementation from~\cite{clark-etal-2021-thats}, and provide detailed instructions and examples to improve evaluation accuracy. 
For this human evaluation, there are four individual annotators in total. We arrange different annotators for different baselines, and each baseline has two annotators.
The Inter-Annotator Agreement for ChatGPT, LLaMA-7B, and Vicuna-13B are reported as follows: 90\%, 97\%, and 89\% respectively.

\begin{table}[ht]
    \centering
    \small
    \begin{tabular}{>{\raggedright\arraybackslash\tt}p{0.45\textwidth}<{}}
    \hline
        Annotation Method: \\
        \\
        Each evaluation content includes a sentence and a piece of knowledge. Our task is to determine whether this sentence contains the given knowledge, i.e., whether this knowledge provides support for the sentence. If the sentence does not mention the given knowledge or if the content of the sentence does not align with the knowledge, it is considered unsupported. We use 1 to indicate support and 0 to indicate lack of support.\\
        \\
        Here are some examples:\\
        \\
        Sentence: Stephen Crane was an American writer born on November 1, 1871, in Newark, and died on June 5, 1900, in Badenweiler.\\
        Knowledge: date of birth: 1871-11-01\\
        Result: 1, because the sentence's date of birth matches the knowledge's date of birth.\\
        \\
        Sentence: Merton died on December 10, 1968, in Bangkok, Thailand.\\
        Knowledge: country of citizenship: United States of America\\
        Result: 0, because the sentence does not mention Merton's nationality.\\
    \hline
  \end{tabular}
  \caption{Instruction we provide to the human annotators. }
  \label{tab:human_eval_template}
\end{table}

\section{Prompts} \label{appendix - prompts}
We present the prompts and instructions we used in this section. We present the prompts for the evolutionary question construction in Table \ref{tab:question_initial_prompt}, \ref{tab:question_other_prompt}, \ref{tab:question_general_prompt}, and \ref{tab:question_specific_prompt}. We present the prompt for the answer generation in Table \ref{tab:main_prompt}. We present the prompts we use for text evaluation with G-Eval in Table \ref{tab:gpt_eval_coherence}, \ref{tab:gpt_eval_consistency}, \ref{tab:gpt_eval_fluency}, and \ref{tab:gpt_eval_relevance}.

\begin{table*}[t]
    \centering
    \begin{tabular}{>{\raggedright\arraybackslash\tt}p{0.97\textwidth}<{}}
    \hline
        Instruction: Your objective is to select relevant knowledge to label the sentence and generate a question \\
        \\
        sentence: Artemisia Gentileschi was born Artemisia Gentileschi Lomi in Rome on July 8 1593 although her birth certificate from the Archivio di Stato indicated she was born in 1590 the eldest child of the Tuscan painter Orazio Gentileschi and Prudenzia di Ottaviano Montoni.
        \\
        knowledge: {\color{brown}\{qid: Q367360, name: Orazio Gentileschi, sex or gender: male, place of birth: Pisa, place of death: London, instance of: human, occupation: painter, child: Artemisia Gentileschi, described by source: The Great Theatre of Dutch Painters, notable works: Diana the Huntress, given name: Orazio, topic's main category: Category:Orazio Gentileschi, surname: Gentileschi, genre: portrait, languages spoken: Italian, movement: mannerism, work location: Rome, ethnic group: Italians, date of birth: 1563-07-19, date of death: 1639-02-07\}}\\
        \vspace{-1em}
        {\color{brown}\{qid: Q212657, name: Artemisia Gentileschi, sex or gender: female, place of birth: Rome, place of death: Naples, instance of: human, occupation: painter, member of: Accademia delle Arti del Disegno, father: Orazio Gentileschi, described by source: The Great Theatre of Dutch Painters, notable works: Judith Slaying Holofernes, topic's main category: Category:Artemisia Gentileschi, movement: Caravaggisti, ethnic group: Italians, work location: Florence, depicted by: Artemisia, field of work: painting, surname: Gentileschi, genre: portrait, languages spoken: Italian, position held: court painter, student of: Orazio Gentileschi, spouse: Pierantonio Stiattesi, given name: Artemisia, mother: Prudenzia di Ottaviano Montoni, date of birth: 1596-07-08, date of death: 1654-01-01\}}\\
        \\
        Generated Answer: {\color{blue} Artemisia Gentileschi [qid: Q212657, name: Artemisia Gentileschi] was born Artemisia Gentileschi Lomi in Rome [qid: Q212657, place of birth: Rome] on July 8 1593 [qid: Q212657, date of birth: 1596-07-08] although her birth certificate from the Archivio di Stato indicated she was born in 1590 the eldest child of the Tuscan painter Orazio Gentileschi [qid: Q212657, father: Orazio Gentileschi] [qid: Q367360, name: Orazio Gentileschi, occupation: painter] and Prudenzia di Ottaviano Montoni.}\\\\\hline
  \end{tabular}
  \caption{Instruction and demonstration for initial round of evolutionary question construction. We use brown color for the knowledge, and blue color for the demonstration of expected output.}
  \label{tab:question_initial_prompt}
\end{table*}

\begin{table*}[t]
    \centering
    \begin{tabular}{>{\raggedright\arraybackslash\tt}p{0.97\textwidth}<{}}
    \hline
        Instruction: Your objective is to extend the original paragraph by adding one sentence that includes the given knowledge\\
        \\
        answer: Artemisia Gentileschi [qid: Q212657, name: Artemisia Gentileschi] was born Artemisia Gentileschi Lomi in Rome [qid: Q212657, place of birth: Rome] on July 8 1593 [qid: Q212657, date of birth: 1596-07-08] although her birth certificate from the Archivio di Stato indicated she was born in 1590 the eldest child of the Tuscan painter Orazio Gentileschi [qid: Q212657, father: Orazio Gentileschi] [qid: Q367360, name: Orazio Gentileschi, occupation: painter] and Prudenzia di Ottaviano Montoni.\\
        knowledge: {\color{brown}\{qid: Q212657, name: Artemisia Gentileschi, notable works: Judith Slaying Holofernes\}}\\
        \\
        Generated Answer: {\color{blue} Artemisia Gentileschi [qid: Q212657, name: Artemisia Gentileschi] was born Artemisia Gentileschi Lomi in Rome [qid: Q212657, place of birth: Rome] on July 8 1593 [qid: Q212657, date of birth: 1596-07-08] although her birth certificate from the Archivio di Stato indicated she was born in 1590 the eldest child of the Tuscan painter Orazio Gentileschi [qid: Q212657, father: Orazio Gentileschi] [qid: Q367360, name: Orazio Gentileschi, occupation: painter] and Prudenzia di Ottaviano Montoni. Under the influence of her father, Artemisia Gentileschi created her iconic painting Judith Slaying Holofernes [qid: Q212657, notable works: Judith Slaying Holofernes] when she was around twenty years old.}\\\hline
  \end{tabular}
  \caption{Instruction and demonstration for evolutionary question construction for round 2 to 5. We use brown color for the knowledge, and blue color for the demonstration of expected output.}
  \label{tab:question_other_prompt}
\end{table*}

\begin{table*}[t]
    \centering
    \begin{tabular}{>{\raggedright\arraybackslash\tt}p{0.97\textwidth}<{}}
    \hline
        Instruction: Your objective is to ask a question whose answer is the given paragraph. There should be only one question when possible, if not, make sure the question is as concise as possible.\\
        \\
        Paragraph: Artemisia Gentileschi was born Artemisia Gentileschi Lomi in Rome on July 8 1593 although her birth certificate from the Archivio di Stato indicated she was born in 1590 the eldest child of the Tuscan painter Orazio Gentileschi and Prudenzia di Ottaviano Montoni. Her life and work were later depicted in the film \"Artemisia\", which brought her story to a wider audience. Her father, Orazio, was a prominent figure in the Mannerism art movement, which likely influenced Artemisia's own artistic style. However, Artemisia herself was a part of the Caravaggisti movement, a group of artists who followed the style of Caravaggio. She was also a student of her father, Orazio Gentileschi, which further shaped her artistic development. Orazio’s influence on Artemisia’s development as a prominent Baroque painter can be seen in her highly naturalistic portrayal of figures, dramatic scenes and the use of chiaroscuro technique\\
        \\
        Generated Question: {\color{blue} Who was Artemisia Gentileschi and what influences shaped her artistic style?}\\\hline
  \end{tabular}
  \caption{Instruction and demonstration for general question generation. We use blue color for the demonstration of expected output.}
  \label{tab:question_general_prompt}
\end{table*}

\begin{table*}[t]
    \centering
    \begin{tabular}{>{\raggedright\arraybackslash\tt}p{0.97\textwidth}<{}}
    \hline
        Instruction: Your objective is to ask a question whose answer is the given paragraph. The question should not be too tedious.\\
        \\
        Paragraph: Artemisia Gentileschi was born Artemisia Gentileschi Lomi in Rome on July 8 1593 although her birth certificate from the Archivio di Stato indicated she was born in 1590 the eldest child of the Tuscan painter Orazio Gentileschi and Prudenzia di Ottaviano Montoni. Her life and work were later depicted in the film \"Artemisia\", which brought her story to a wider audience. Her father, Orazio, was a prominent figure in the Mannerism art movement, which likely influenced Artemisia's own artistic style. However, Artemisia herself was a part of the Caravaggisti movement, a group of artists who followed the style of Caravaggio. She was also a student of her father, Orazio Gentileschi, which further shaped her artistic development. Orazio’s influence on Artemisia’s development as a prominent Baroque painter can be seen in her highly naturalistic portrayal of figures, dramatic scenes and the use of chiaroscuro technique\\
        \\
        Generated Question: {\color{blue} What were the key artistic influences and characteristics that shaped Artemisia Gentileschi's unique Baroque style, and how did her relationship with her father, Orazio Gentileschi, impact her artistic development?}\\\hline
  \end{tabular}
  \caption{Instruction and demonstration for specific question generation. We use blue color for the demonstration of expected output.}
  \label{tab:question_specific_prompt}
\end{table*}

\begin{table*}[t]
    \centering
    \begin{tabular}{>{\raggedright\arraybackslash\tt}p{0.97\textwidth}<{}}
    \hline
        Instruction: You answer the question based on your knowledge, with the given information for annotation, following the given format. Use [NA] for claims that need annotation but is unprovided.\\
        \\
        Question: Considering the information:\\
        \vspace{-1em}
        {\color{brown}\{name: Orazio Gentileschi, place of death: London, child: Artemisia Gentileschi, notable works: Diana the Huntress, given name: Orazio, surname: Gentileschi, languages spoken: Italian, movement: mannerism, work location: Rome, ethnic group: Italians, date of birth: 1563-07-19, date of death: 1639-02-07, qid: Q367360\}}\\
        \vspace{-1em}
        {\color{brown}\{name: Artemisia Gentileschi, place of birth: Rome, place of death: Naples, occupation: painter, member of: Accademia delle Arti del Disegno, father: Orazio Gentileschi, notable works: Judith Slaying Holofernes, movement: Caravaggisti, ethnic group: Italians, work location: Florence, depicted by: Artemisia, surname: Gentileschi, languages spoken: Italian, student of: Orazio Gentileschi, given name: Artemisia, mother: Prudenzia di Ottaviano Montoni, date of death: 1654-01-01, qid: Q212657\}}\\
        How did Orazio Gentileschi's influence on Artemisia's life and career shape her development as a prominent Baroque painter, despite facing significant challenges as a female artist in a male-dominated field?\\
        \\
        Answer: {\color{blue} Artemisia Gentileschi was an Italian painter born on July 8, 1596 [NA] in Rome [Q212657, ethnic group: Italians, occupation: painter, place of birth: Rome]. She was a member of the Accademia delle Arti del Disegno and is best known for her work Judith Slaying Holofernes [Q212657, member of: Accademia delle Arti del Disegno, notable works: Judith Slaying Holofernes]. She was the eldest child of the Tuscan painter Orazio Gentileschi [Q212657, father: Orazio Gentileschi]. Orazio Gentileschi was an Italian painter [NA] born in 1563 and died in 1639 [Q367360, ethnic group: Italians, date of birth: 1563-07-19, date of death: 1639-02-07]. He was born in Pisa and died in London [Q367360, place of death: London]. Orazio’s influence on Artemisia’s development as a prominent Baroque [NA] painter can be seen in her highly naturalistic portrayal of figures, dramatic scenes and the use of chiaroscuro technique [NA]. He also provided her with the opportunity to study with him and learn from his experience and expertise. She became an important second-generation proponent of Caravaggio’s dramatic realism [Q212657, movement: Caravaggisti].}\\\hline
  \end{tabular}
  \caption{Full instruction and demonstration for answer generation with citaion. We use brown color for the knowledge pool, and blue color for the expected outcome provided by the demonstration.}
  \label{tab:main_prompt}
\end{table*}

\begin{table*}[t]
    \centering
    \begin{tabular}{>{\raggedright\arraybackslash\tt}p{0.97\textwidth}<{}}
    \hline
        Instruction: You will be given one question and answer. Your task is to rate the answer on one metric. Please make sure you read and understand these instructions carefully. Please keep this document open while reviewing, and refer to it as needed. \\
        \\Evaluation Criteria:\\Coherence (1-5) - the collective quality of all sentences. We align this dimension with the DUC quality question of structure and coherence whereby the answer should be well-structured and well-organized. The answer should not just be a heap of related information, but should build from sentence to sentence to a coherent body of information about a topic.\\
        \\Evaluation Steps:\\1. Read the questions carefully and identify the main topic and key points.\\2. Read the answer and compare it to the question. Check if the answer covers the main topic and key points of the question, and if it presents them in a clear and logical order.\\3. Assign a score for coherence on a scale of 1 to 5, where 1 is the lowest and 5 is the highest based on the Evaluation Criteria.\\\hline
  \end{tabular}
  \caption{Instruction for text evaluation with GPT-EVAL - Coherence}
  \label{tab:gpt_eval_coherence}
\end{table*}

\begin{table*}[t]
    \centering
    \begin{tabular}{>{\raggedright\arraybackslash\tt}p{0.97\textwidth}<{}}
    \hline
        Instruction: You will be given one question and answer. Your task is to rate the answer on one metric. Please make sure you read and understand these instructions carefully. Please keep this document open while reviewing, and refer to it as needed. \\
        \\Evaluation Criteria:\\Consistency (1-5) - the answer should be consistent with the given knowledge. The answer should also be self-consistent, without any contradiction to itself.\\
        \\
        Evaluation Steps:\\1. Read the question and knowledge carefully.\\2. Read the answer and compare it to the knowledge. Check if the answer is consistent with the give knowledge.\\3. Assign a score for consistency on a scale of 1 to 5, where 1 is the lowest and 5 is the highest based on the Evaluation Criteria.\\\hline
  \end{tabular}
  \caption{Instruction for text evaluation with GPT-EVAL - Consistency}
  \label{tab:gpt_eval_consistency}
\end{table*}

\begin{table*}[t]
    \centering
    \begin{tabular}{>{\raggedright\arraybackslash\tt}p{0.97\textwidth}<{}}
    \hline
        Instruction: You will be given one question and answer. Your task is to rate the answer on one metric. Please make sure you read and understand these instructions carefully. Please keep this document open while reviewing, and refer to it as needed. \\
        \\Evaluation Criteria:\\Fluency (1-5) - the answer should be written in fluent language. The answer should use appropriate vocabulary, grammar, and sentence structures that enable readers or listeners to comprehend the content effortlessly.\\
        \\
        Evaluation Steps:\\1. Read the question carefully.\\2. Read the answer and check if the language in the answer is fluent.\\3. Assign a score for fluency on a scale of 1 to 5, where 1 is the lowest and 5 is the highest based on the Evaluation Criteria.\\\hline
  \end{tabular}
  \caption{Instruction for text evaluation with GPT-EVAL - Fluency}
  \label{tab:gpt_eval_fluency}
\end{table*}

\begin{table*}[t]
    \centering
    \begin{tabular}{>{\raggedright\arraybackslash\tt}p{0.97\textwidth}<{}}
    \hline
        Instruction: You will be given one question and answer. Your task is to rate the answer on one metric. Please make sure you read and understand these instructions carefully. Please keep this document open while reviewing, and refer to it as needed. \\
        \\Evaluation Criteria:\\Relevance (1-5) - the answer should be relevant to the question. The answer should directly answers the question, without providing any irrelevant information.\\
        \\
        Evaluation Steps:\\1. Read the question carefully.\\2. Read the answer and compare with the question to check if it fully answers the question and have no redundancies.\\3. Assign a score for relevance on a scale of 1 to 5, where 1 is the lowest and 5 is the highest based on the Evaluation Criteria.\\\hline
  \end{tabular}
  \caption{Instruction for text evaluation with GPT-EVAL - Relevance}
  \label{tab:gpt_eval_relevance}
\end{table*}

\section{Evolutionary Question Generation} \label{appendix - evolutionary}
We provide an example of evolutionary question generation in Table \ref{tab:evolutionary}.

\begin{table*}[t]
    \centering
    \small
    \begin{tabular}{>{\raggedright\arraybackslash\tt}p{0.97\textwidth}<{}}
    \hline
        Round 1:\\
        \textbf{Annotated Knowledge}: \\
        \vspace{-1em}
        {\color{brown}[qid: Q258115, name: Diego Simeone, occupation: association football player]}\\
        \vspace{-1em}
        {\color{brown}[qid: Q6439494, name: Giovanni Simeone, father: Diego Simeone]}\\
        \vspace{-1em}
        {\color{brown}[qid: Q6439494, name: Giovanni Simeone, place of birth: Madrid]}\\
        \vspace{-1em}
        {\color{brown}[qid: Q258115, name: Diego Simeone, member of sports team: Atletico Madrid]}\\
        \vspace{-1em}
        \textbf{Paragraph}: The son of former Argentine international footballer Diego Simeone and Carolina Baldini Giovanni Simeone was born in Madrid while his father was playing for Atletico Madrid. \\
        \\
    \hline
        Round 2:\\
        \textbf{Input Knowledge}: {\color{brown}[qid: Q258115, name: Diego Simeone, position played on team: midfielder]}\\
        \textbf{Paragraph}: The son of former Argentine international footballer Diego Simeone and Carolina Baldini Giovanni Simeone was born in Madrid while his father was playing \underline{as a midfielder} for Atletico Madrid. \\
        \\
    \hline
        Round 3:\\
        \textbf{Input Knowledge}: {\color{brown}[qid: Q258115, Diego Simeone, member of sports team: S.S. Lazio]}\\
        \textbf{Paragraph}: The son of former Argentine international footballer Diego Simeone and Carolina Baldini Giovanni Simeone was born in Madrid while his father was playing as a midfielder for Atletico Madrid. Before his stint at Atletico Madrid, \underline{Diego Simeone had also played for S.S. Lazio.} \\
        \\
    \hline
        Round 4:\\
        \textbf{Input Knowledge}: {\color{brown}[qid: Q258115, Diego Simeone, medical condition: COVID-19]}\\
        \textbf{Paragraph}: The son of former Argentine international footballer Diego Simeone and Carolina Baldini Giovanni Simeone was born in Madrid while his father was playing as a midfielder for Atletico Madrid. Before his stint at Atletico Madrid, Diego Simeone had also played for S.S. Lazio. Later in life, \underline{Diego Simeone contracted the} \underline{COVID-19 virus,} but managed to recover. \\
        \\
    \hline
        Round 5:\\
        \textbf{Input Knowledge}: {\color{brown}[qid: Q6439494, Giovanni Simeone, position played on team: forward]}\\
        \textbf{Paragraph}: The son of former Argentine international footballer Diego Simeone and Carolina Baldini Giovanni Simeone was born in Madrid while his father was playing as a midfielder for Atletico Madrid. Before his stint at Atletico Madrid, Diego Simeone had also played for S.S. Lazio. Later in life, Diego Simeone contracted the COVID-19 virus, but managed to recover. Following in his father's footsteps but with a different role on the pitch, \underline{Giovanni Simeone grew up to be a forward in the game of football.} \\
        \\
    \hline
        \textbf{General Question}:\\
        \vspace{-1em}
        {\color{blue}How did the career of Diego Simeone influence his son's path in football?}\\
        \textbf{Specific Question}:\\
        \vspace{-1em}
        {\color{blue} Who are Diego Simeone and Giovanni Simeone, and how are there career path in football different from each other?}\\
    \hline
  \end{tabular}
  \caption{An example of the generated questions after each round in evolutionary question construction. We use brown color for the knowledge, and blue color for the generated questions. In each round we underline the texts to the newly injected knowledge.}
  \label{tab:evolutionary}
\end{table*}

\section{Examples} \label{appendix - examples}
We show examples of the attributed answers generated by the LLMs in Table \ref{tab:example_chatgpt} and \ref{tab:example_gpt4}.

\begin{table*}[t]
    \centering
    \begin{tabular}{>{\raggedright\arraybackslash\tt}p{0.97\textwidth}<{}}
    \hline
        \textbf{Question}:\\
        \vspace{-1em}
        {\color{blue}What were some of Stephen Crane's notable works that reflected his views on religion and the harsh realities of life, and how did his experiences at Syracuse University influence his writing? Also, what was the name of the sports team he was a member of at Syracuse University?}\\
        \\
        \textbf{Retrieved Knowledge}: \\
        \vspace{-1em}
        {\color{brown}[qid: Q206534, name: Stephen Crane, sex or gender: male, place of birth: Newark, place of death: Badenweiler, country of citizenship: United States of America, cause of death: Tuberculosis, place of burial: Evergreen Cemetery, alma mater: Syracuse University, instance of: human, given name: Stephen, medical condition: Tuberculosis, occupation: writer, described by source: Great Soviet Encyclopedia (1969â€“1978), manner of death: natural causes, movement: literary realism, notable works: The Red Badge of Courage, surname: Crane', "topic's main category": 'Category:Stephen Crane, religion: atheism, languages spoken: English, mother: Mary Helen Peck Crane, sport: baseball, member of sports team: Syracuse Orange baseball, father: Jonathan Townley Crane, name: Stephen Crane, date of birth: 1871-11-01, date of death: 1900-06-05]}\\
        \\
        \textbf{Generated Output}: \\
        \vspace{-1em}
        Stephen Crane was a male writer born on November 1, 1871, in Newark, United States of America, and died on June 5, 1900, in Badenweiler due to tuberculosis [Q206534, sex or gender: male, place of birth: Newark, place of death: Badenweiler, country of citizenship: United States of America, cause of death: Tuberculosis, medical condition: Tuberculosis, date of birth: 1871-11-01, date of death: 1900-06-05]. Crane was a prominent literary realist writer [Q206534, movement: literary realism] who wrote about the harsh realities of life and his atheistic views on religion [Q206534, religion: atheism]. His notable works include "The Red Badge of Courage" [Q206534, notable works: The Red Badge of Courage]. \\
        \\
        Crane attended Syracuse University and was a member of the Syracuse Orange baseball team [Q206534, alma mater: Syracuse University, sport: baseball, member of sports team: Syracuse Orange baseball]. His experiences at Syracuse University influenced his writing, and he wrote several stories about college life, including "The Eighty-Yard Run," which was based on his own experiences as a football player at Syracuse [NA].\\
    \hline
  \end{tabular}
  \caption{An example of the attributed answer generated by ChatGPT (temperature 0.5). We use blue color for the question and brown color for the retrieved knowledge.}
  \label{tab:example_chatgpt}
\end{table*}

\begin{table*}[t]
    \centering
    \begin{tabular}{>{\raggedright\arraybackslash\tt}p{0.97\textwidth}<{}}
    \hline
        \textbf{Question}:\\
        \vspace{-1em}
        {\color{blue}What were some of Stephen Crane's notable works that reflected his views on religion and the harsh realities of life, and how did his experiences at Syracuse University influence his writing? Also, what was the name of the sports team he was a member of at Syracuse University?}\\
        \\
        \textbf{Retrieved Knowledge}:\\
        \vspace{-1em}
        {\color{brown}[qid: Q206534, name: Stephen Crane, sex or gender: male, place of birth: Newark, place of death: Badenweiler, country of citizenship: United States of America, cause of death: Tuberculosis, place of burial: Evergreen Cemetery, alma mater: Syracuse University, instance of: human, given name: Stephen, medical condition: Tuberculosis, occupation: writer, described by source: Great Soviet Encyclopedia (1969â€“1978), manner of death: natural causes, movement: literary realism, notable works: The Red Badge of Courage, surname: Crane', "topic's main category": 'Category:Stephen Crane, religion: atheism, languages spoken: English, mother: Mary Helen Peck Crane, sport: baseball, member of sports team: Syracuse Orange baseball, father: Jonathan Townley Crane, name: Stephen Crane, date of birth: 1871-11-01, date of death: 1900-06-05]}\\
        \\
        \textbf{Generated Output}:\\
        \vspace{-1em}
        Stephen Crane was an American writer born on November 1, 1871, in Newark, and died on June 5, 1900, in Badenweiler [Q206534, date of birth: 1871-11-01, date of death: 1900-06-05, place of birth: Newark, place of death: Badenweiler]. He was a part of the literary realism movement and is best known for his notable work, The Red Badge of Courage [Q206534, movement: literary realism, notable works: The Red Badge of Courage]. Crane's views on religion were atheistic [Q206534, religion: atheism]. His works often reflected the harsh realities of life, such as war, poverty, and human suffering [NA].\\
        \\
        Crane attended Syracuse University [Q206534, alma mater: Syracuse University], where his experiences likely influenced his writing. He was exposed to a diverse range of ideas and perspectives, which may have contributed to his realistic approach to literature [NA].\\
        \\
        While at Syracuse University, Stephen Crane was a member of the Syracuse Orange baseball team [Q206534, member of sports team: Syracuse Orange baseball]. \\
    \hline
  \end{tabular}
  \caption{An example of the attributed answer generated by GPT-4. We use blue color for the question and brown color for the retrieved knowledge.}
  \label{tab:example_gpt4}
\end{table*}
\label{sec:appendix}

\end{document}